\documentclass{article}
\PassOptionsToPackage{numbers, sort&compress}{natbib}

\usepackage[preprint]{neurips_2024}

\usepackage[utf8]{inputenc} %
\usepackage[T1]{fontenc}    %
\usepackage{hyperref}       %
\usepackage{url}            %
\usepackage{booktabs}       %
\usepackage{amsfonts}       %
\usepackage{nicefrac}       %
\usepackage{microtype}      %
\usepackage{xcolor}         %
\usepackage{graphicx} %
\usepackage{float} %
\usepackage{wrapfig}
\usepackage{subfigure}
\usepackage{colortbl}
\usepackage{multirow}
\usepackage{amsmath}
\usepackage{amssymb}
\usepackage{bbding}

\newcommand{\tabref}[1]{Table \ref{#1}}
\newcommand{\figref}[1]{Figure \ref{#1}}

\def\ie{\emph{i.e.}}
\def\eg{\emph{e.g.}}
\def\etc{\emph{etc}}
\def\etal{{\em et al.~}}
\def\ourmodel{\emph{SqueezeTime}}

\title{No Time to Waste: Squeeze Time into Channel for Mobile Video Understanding}

\author{
	Yingjie Zhai, Wenshuo Li, Yehui Tang, Xinghao Chen$^*$, Yunhe Wang\thanks{Corresponding authors}\\
	\small Huawei Noah's Ark Lab.\\
	\small\texttt{\{xinghao.chen,yunhe.wang\}@huawei.com}\\
}

\begin{document}

	\maketitle

\begin{abstract}
	Current architectures for video understanding mainly build upon 3D convolutional blocks or 2D convolutions with additional operations for temporal modeling.
	However, these methods all regard the temporal axis as a separate dimension of the video sequence, which requires large computation and memory budgets and thus limits their usage on mobile devices.
	In this paper, we propose to squeeze the time axis of a video sequence into the channel dimension and present a lightweight video recognition network, term as \textit{SqueezeTime}, for mobile video understanding.
	To enhance the temporal modeling capability of the proposed network, we design a Channel-Time Learning (CTL) Block to capture temporal dynamics of the sequence.
	This module has two complementary branches, in which one branch is for temporal importance learning and another branch with temporal position restoring capability is to enhance inter-temporal object modeling ability.
	The proposed SqueezeTime is much lightweight and fast with high accuracies for mobile video understanding.
	Extensive experiments on various video recognition and action detection benchmarks, \ie, Kinetics400, Kinetics600, HMDB51, AVA2.1 and THUMOS14, demonstrate the superiority of our model.
	For example, our SqueezeTime achieves $+1.2\%$ accuracy and $+80\%$ GPU throughput gain on Kinetics400 than prior methods.
	Codes are publicly available at 
	\href{https://github.com/xinghaochen/SqueezeTime}{https://github.com/xinghaochen/SqueezeTime} and \href{https://github.com/mindspore-lab/models/tree/master/research/huawei-noah/SqueezeTime}{https://github.com/mindspore-lab/models/tree/master/research/huawei-noah/SqueezeTime}.
\end{abstract}

\section{Introduction}
\vspace{-10pt}
In recent years, the amount of videos is of explosive growth.
Different from image recognition, processing these video data is much more resource-consuming.
For example, when simply extending the image recognition model, \eg, ResNet~\cite{resnet}, into the temporal-spatial version, the multiply-add operations are increased by $27$ times~\cite{x3d}, which is much compute-demanding.
Therefore, research on how to design efficient video understanding models is becoming a hot topic, which has many
applications, \eg, mobile video analysis, autonomous driving, robotics, industrial control, \etc.
\par
\begin{figure}[h]
	\subfigure[]{
		\label{Fig.sub.1}
		\includegraphics[width=0.47\linewidth]{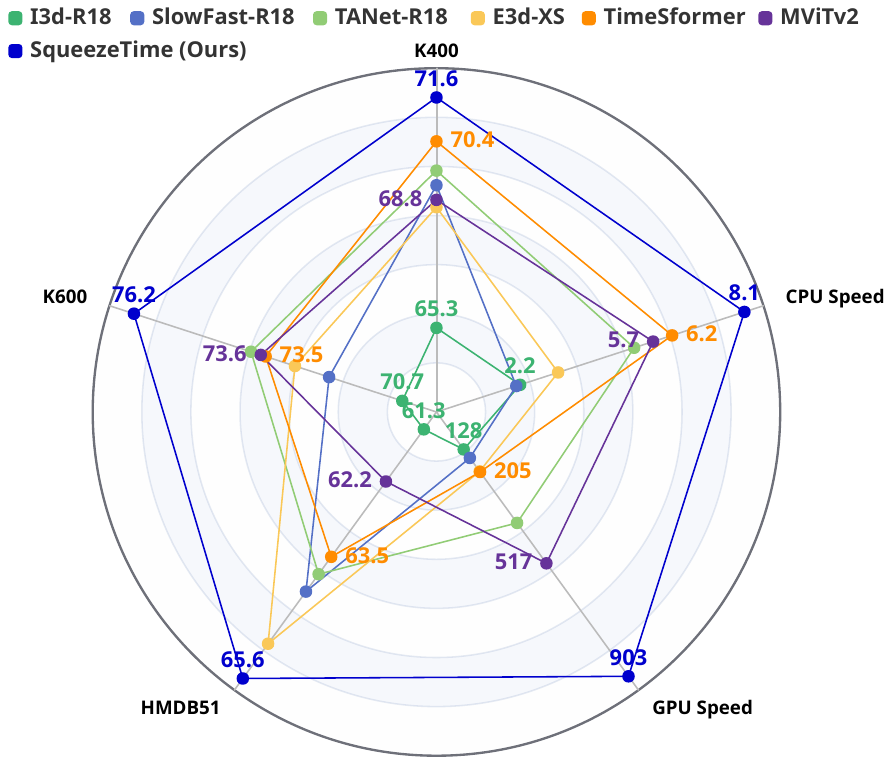}}
	\subfigure[]{
		\label{Fig.sub.2}
		\includegraphics[width=0.47\linewidth]{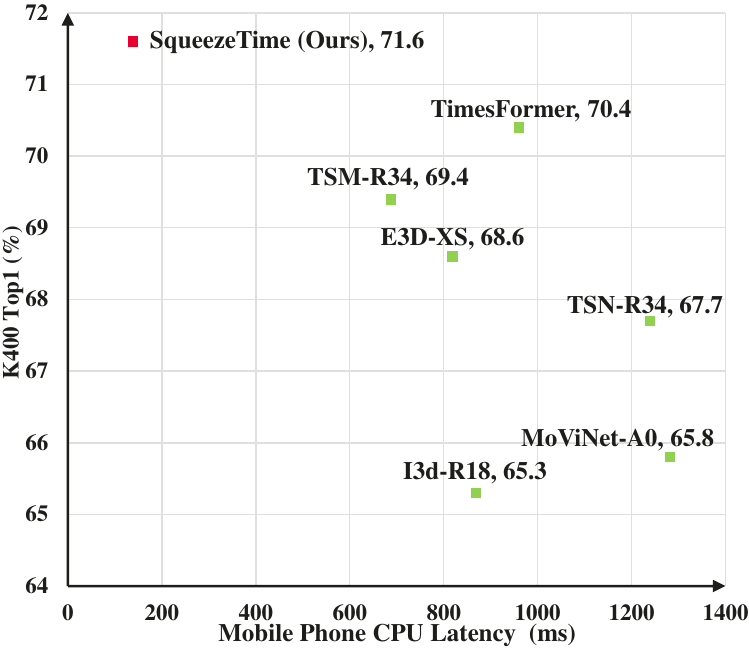}}
	\vspace{-10pt}
	\caption{(a) Performance comparison for mobile video recognition of multiple methods on K400~\cite{k400}, K600~\cite{k600} and HMDB51~\cite{hmdb51} datasets. We report Top1 accuracy ($\%$), GPU Speed (throughput, videos / s), and CPU Speed (videos / s) on the figure. Note the CPU speed is measured by the `1 / latency (ms)' for better visualization. (b) Latency comparison of multiple models on a mobile device. }
	\label{fig:fig1}
	\vspace{-15pt}
\end{figure}

However, it nowadays remains difficult to find models that can run fast on edge devices with high accuracy.  
Traditional 3D convolutional networks, \eg, 3D ConvNets~\cite{c3d}, I3D~\cite{i3d} and SlowFast~\cite{slowfast} can jointly learn spatial and temporal features from videos but consume large amounts of memory and computation, which makes them not suitable for the mobile usage.
To make the 3D convolutional network more efficient, some previous methods tried to improve the 3D convolutional network by 2D decomposition or approximation manually~\cite{res3d,xie2018rethinking,r2plus1d,x3d}.
These manually designed methods cost massive effort and time, thus a series of other methods~\cite{movinets,e3d} used the Neural Architecture Search (NAS) technique to automatically design 3D video recognition architectures.
However, searching for such 3D architectures on a video benchmark is also time-consuming (a lot of days on GPUs or TPUs) or handware-dependent~\cite{e3d}.
Another kind of popular methods introduce the 2D CNN into the video recognition task by incorporating extra temporal learning mechanisms, \eg, Temporal Shift Module~\cite{tsm}, Temporal Difference Moudle~\cite{tdn}, Temporal Adaptive Module~\cite{tam}, Adaptive Focus~\cite{adafocus}, Temporal Patch Shift~\cite{xiang2022spatiotemporal}, Temporally-Adaptive Convolutions~\cite{tada}, \etc.
Though these methods have improved running speed, the accuracies are not quite satisfactory in mobile settings.
Recently, transformers-based models~\cite{timeformer,mvitv2,videoswin,aim,survey_on_pami} are proposed for video analysis with encouraging performance on various datasets, but they are not friendly to mobile devices.
\par
\begin{figure}[t]
	\centering
	\includegraphics[width=0.7\linewidth]{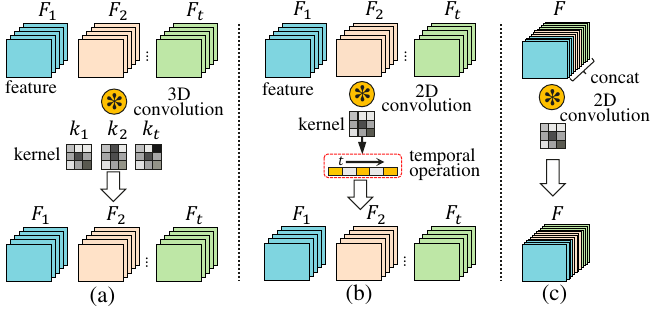}
	\vspace{-10pt}
	\caption{Feature and kernel illustration of different video models, \ie, (a) 3D CNN-based models, (b) 2D CNN with temporal modeling operations, and (c) the proposed squeeze time mechanism.
	}
	\label{fig:cmp}
	\vspace{-15pt}
\end{figure}

Note that the above models treat the temporal axis of videos as an extra dimension, and keep the original temporal dimension when forwarding the video feature (see \figref{fig:cmp}).
Such kinds of operations cost large amounts of memory and need extra computation costs for the extra dimension, which is not efficient enough for mobile computing.
In this paper, we reveal that it is not necessary to keep the temporal axis to build a backbone network for mobile video tasks.
Therefore, we propose to squeeze the temporal axis of the video data into the spatial channel dimension and design a corresponding lightweight backbone, \ie, {\it\textbf{SqueezeTime}}, for mobile video understanding.
It is demonstrated fast with high accuracies on both GPU and (Mobile) CPU devices for video recognition (please see \figref{fig:fig1}).
To eliminate the negative impact caused by the squeezing operation, we propose a Channel-Time Learning Block (CTL) to learn the temporal dynamics embedded into the channels.
Specifically, the CTL block contains two branches: one branch with \textbf{T}emporal \textbf{F}ocus \textbf{C}onvolution (TFC) concentrates on learning the potential temporal importance of different channels, and another branch is leveraged to restore the temporal information of multiple channels and to model the \textbf{I}nter-temporal \textbf{O}bject \textbf{I}nteraction (IOI) using large kernels.
Built with these CTL blocks, the proposed \ourmodel~can capture strong video representations using only a 2D CNN with no extra resource consumption brought by the temporal dimension, which is much more efficient for mobile video analysis.
Note most previous work concentrates on designing models with higher performance using large computation costs ($>40$ GFLOPs), which is not suitable for comparing mobile video recognition.
Thus, in this work, we also make experiments to evaluate the lightweight version of popular models for mobile video analysis, which demonstrates the proposed~\ourmodel~can exceed other state-of-the-art methods on various datasets.
\par
In summary, the key contributions of our work are:
\vspace{-5pt}
\begin{itemize}
	\item We propose to squeeze the temporal dimension of the video sequence into spatial channels, which is much faster with low memory-consuming and low computation cost.
	\vspace{-2pt}
	\item We elaborately design the~\ourmodel~with CTL blocks. The CTL can learn the potential temporal importance of channels, restore temporal information, and enhance inter-temporal object modeling ability, which brings $4.4\%$ Top1 accuracy gain on K400.
	\vspace{-2pt}
	\item We thoroughly benchmark popular video models in the mobile video analysis settings. Extensive experiments demonstrate the proposed \ourmodel~can yield higher accuracy ($+1.2\%$ Top1 on K400) with faster CPU and GPU ($+80\%$ throughput on K400) speed. 
\end{itemize}
\vspace{-15pt}
\section{Related Work}
\vspace{-10pt}
\textbf{CNN-based Methods.} For video recognition models, it is important to capture temporal relations between video frames~\cite{robust}. CNN-based methods can be divided into two categories to address the problem~\cite{kong2022human}.
One kind of method intuitively designs efficient video recognition networks by stacking 3D convolution~\cite{channelseparated,c3d,i3d,x3d,slowfast,e3d,movinets}, which is used to directly extract spatial-temporal representations from video clips~\cite{res3d, channelseparated}.
For example, X3D~\cite{x3d} proposed a family of 3D CNNs by expanding a tiny 2D image classifier along multiple network axes, \ie, space, time, width, and depth.
MoViNet~\cite{movinets} designed a three-step approach to improve computational efficiency while reducing the peak memory usage of 3D CNNs.
E3D~\cite{e3d} derived an analytic entropy searching strategy to automatically design 3D CNN architectures.
Another kind of method designs efficient video recognition models by equipping the 2D CNNs with the temporal modeling capacity~\cite{tpn, nonlocal,r2plus1d,tsm,moreisless,tin,tpn,tsn,tdn,tea,teinet,tam, actionet, tada}.
For example, Lin \etal\cite{tsm} proposed a generic and effective temporal shift module (TSM), which can be inserted into 2D CNNs at zero computation and zero parameters.
Liu \etal\cite{tam} presented a novel temporal adaptive module (TAM) to produce video-specific temporal kernels based on feature maps to capture diverse motion patterns.
Huang \etal\cite{tada} designed a temporal-adaptive convolution (TAdaConv) for video understanding.
\par

\noindent\textbf{Transformer-based Methods. }
Recently, vision transformers are introduced to video understanding~\cite{vivit,zhao2022alignment,mvitv2,yan2022multiview,yang2022recurring,videoswin,xiang2022spatiotemporal,fan-iclr2022,timeformer,aim,wu2023bidirectional}.
For example, ViViT~\cite{vivit} presented pure-transformer models by factorizing the model along spatial and temporal dimensions to increase efficiency and scalability.
Video Swin Transformer~\cite{videoswin} designed an inductive bias of locality in video transformers to get a better speed-accuracy trade-off.
MViTv2~\cite{mvitv2} incorporated decomposed relative positional embeddings and residual pooling connections into the transformer for video recognition.
TimeSformer~\cite{timeformer} suggested divided attention, which separately applies temporal attention and spatial attention with each block, leading to good performance.
Although transformer-based methods achieve strong performance, they are not friendly to mobile usage.
\par

\noindent\textbf{Mobile Video Understanding.}
Most recent work is focused on designing efficient video recognition models to achieve higher accuracy on various benchmarks. 
But there are few works~\cite {x3d,e3d,movinets} that are devoted to designing lightweight models deployed on mobile devices.
The recent work X3D~\cite{x3d}, MoViNet~\cite{movinets}, and E3D~\cite{e3d} based on stepwise network expansion or neural architecture searching designed several lightweight models, but the trade-off between the speed and accuracy is not satisfactory enough.
Therefore, in this paper, we aim to build a lightweight video backbone with large GPU throughput, low CPU latency, and high accuracy for mobile video understanding.
\vspace{-10pt}
\section{SqueezeTime Networks}
\vspace{-10pt}
\begin{figure*}[t]
	\centering
	\includegraphics[width=0.9\linewidth]{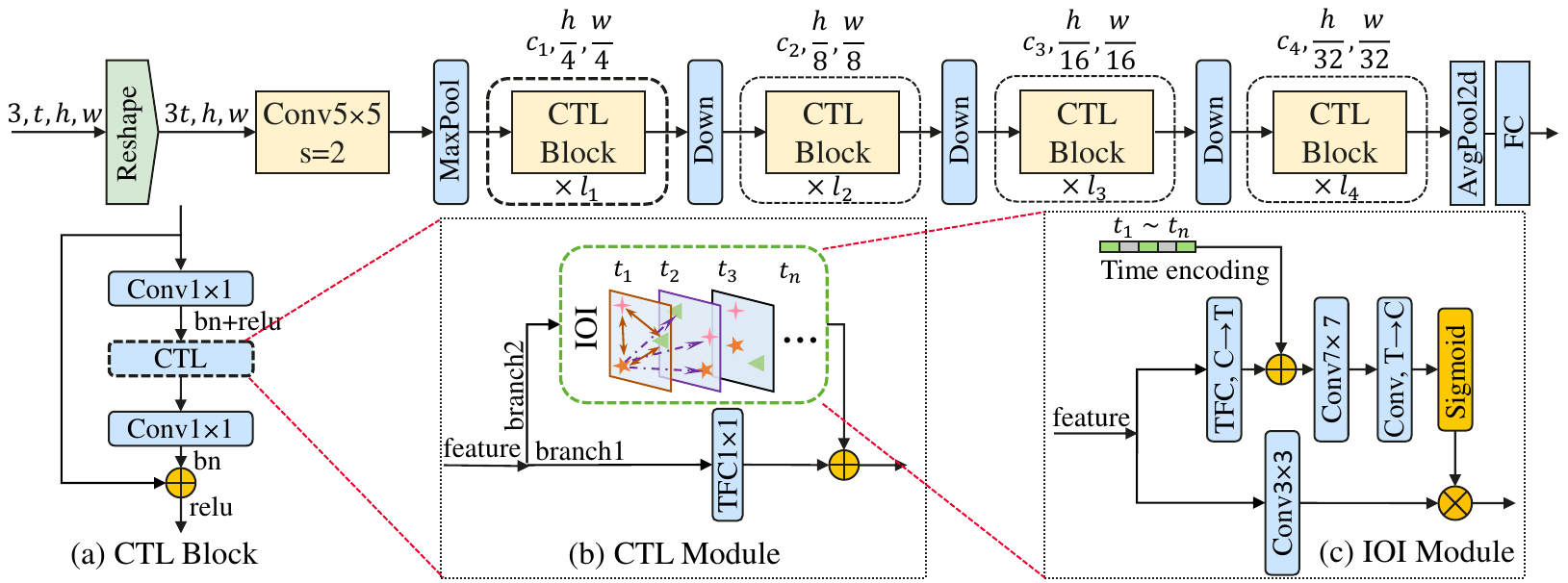}
	\vspace{-5pt}
	\caption{Pipeline of the proposed \ourmodel. The input video clip is first reshaped by squeezing the temporal dimension into channels and is then fed into the following network.
		The proposed network contains four main stages, each stage with a stack of CTL Blocks, which are elaborately designed to excavate and restore hidden temporal representations.
	}
	\label{fig:pipe}
	\vspace{-15pt}
\end{figure*}
The core concept of our work, to design an efficient mobile recognition backbone, is first squeezing the temporal information into channels, which can save a lot of computing resources and memory consumption from the temporal dimension.
Then we need to carefully design the model so that it can excavate important temporal dynamics from these fused channels.
The framework of the proposed model is shown in \figref{fig:pipe}.
Let $X\in R^{3, t, h, w}$ denotes the input video sequence, we first squeeze the temporal axis into channels by the reshape operation and get the squeezed video input $X^{\prime} \in R^{3\times t, h, w}$.
Then we feed it into the stem layer with $5\times 5$ convolution to get the base feature $\mathbf{F}_b$, which is then processed by four stages of blocks to learn temporal dynamics and object representations.
Finally, an average pooling layer followed by a fully-connected layer is used to predict the result.  
At each stage, we use a $2\times2$ convolution with a stride $2$ to downsample larger-scale features.
We will describe the proposed method in detail in the following section.
\vspace{-10pt}
\subsection{Squeeze and Restore Time}
\vspace{-5pt}
Previous CNN-based methods treat the temporal axis of the video sequence as an extra dimension, \eg, 3D CNN-based methods~\cite{i3d,x3d,e3d} and 2D CNN with temporal learning modules~\cite{tsm,tam,tada}.
Let $t, h, w, c_{in}, c_{out}, k$ represent the temporal size, spatial sizes, input channels, output channels, and kernel size of the input feature. The computation complexity of the 3D CNN-based methods and 2D CNN-based methods with temporal learning module are $2c_{out}c_{in}k^3hwt$ and $2c_{out}c_{in}k^2hwt+O(t)$, respectively.
$O(t)$ represents the computation of temporal modeling.
As shown in \figref{fig:cmp}, we propose to squeeze the temporal dimension into spatial channels, which can further reduce the computation complexity to $2c_{out}c_{in}k^2hw$, and more importantly, it can save a lot of memory originally occupied by keeping the temporal axis (please see \figref{fig:cmp} to compare with other models).
The proposed squeeze mechanism is formulated by:
\begin{equation}
	\setlength{\abovedisplayskip}{1pt}
	\setlength{\belowdisplayskip}{1pt}
	\mathbf{F}_b=f_m(f_s(X)),
\end{equation}
where $f_s$ is the squeeze function, $f_m$ is the mix up function, and $\mathbf{F}_b$ is the squeezed feature without temporal dimension.
Note that the temporal information of the squeezed feature $\mathbf{F}_b$ is mixed up, which impedes the learning of discriminate features.
For one thing, the temporal information is out of order.
For another thing, the spatial objects from the same frame may be distributed in different temporal channels.
Therefore, we propose to design a 2D block with temporal importance learning and inter-temporal object interaction function to remedy the problems.
The restoration process is formulated as:
\begin{equation}
	\setlength{\abovedisplayskip}{2pt}
	\setlength{\belowdisplayskip}{2pt}
	\mathbf{F}^{\prime}=\beta(\mathbf{F}_b)+\xi(\mathbf{F}_b+\tau),
	\label{equ:restore}
\end{equation}
where $\beta$ is the temporal importance learning function, $\xi$ is the inter-temporal interaction function, and $\tau$ is the injected temporal order information. $\mathbf{F}^{\prime}$ is the restored feature.
\vspace{-10pt}
\subsection{Channel-Time Learning (CTL) Block}
\vspace{-5pt}
Channel-Time Learning (CTL) Block is the basic component of~\ourmodel~to implement the Formula \ref{equ:restore}. As shown in \figref{fig:pipe} (a), the Channel-Time Learning (CTL) Block, built with CTL Module, follows the bottleneck form of ResNet~\cite{resnet}.
It contains a $1\times 1$ convolution to reduce the channels, a CTL module to learn temporal and spatial representations, and another $1\times 1$ convolution to restore the channel number.
The CTL block can be formulated by:
\begin{equation}
	\setlength{\abovedisplayskip}{2pt}
	\setlength{\belowdisplayskip}{2pt}
	\mathbf{F}_{o}={\rm Conv}_{1\times 1}^{rC\rightarrow C}\left({\rm CTL}\left({\rm Conv}_{1\times 1}^{C\rightarrow rC}(\mathbf{F}_{i})\right)\right)+\mathbf{F}_i,
\end{equation}
where $\mathbf{F}_i$ and $\mathbf{F}_o$ are the input feature and output feature of the CTL block, and $r$ is the ratio controlling the channel expansion.
$\rm CTL(\cdot)$ represents the CTL Module. Note we omit the batch normalization and ReLU operations in the formula.
We set the reduction factor $r$ to $0.25$ as the default.
\par
As shown in \figref{fig:pipe} (a), the channel-time learning module contains two complementary branches.
The bottom branch1 is a temporal focus convolution (TFC) with $1\times 1$ kernel size to especially concentrate on capturing the temporal-channel importance.
The top branch2 is an inter-temporal object interaction (IOI) module that aims to restore the temporal position information and model the inter-channel spatial relations using large kernels.
The final output of the CTL module is the summation of the two branches.
\par
\noindent\textbf{Temporal Focus Convolution (TFC).} When squeezing the temporal dimension into the channels, a natural question is: whether the original 2D convolution is suitable to model the temporal representations hidden in different channels.
For a common 2D convolution, it multiplies and accumulates the values of all channels in a local window size:
\begin{equation}
	\setlength{\abovedisplayskip}{2pt}
	\setlength{\belowdisplayskip}{2pt}
	f(x,y)=\sum_{m=0}^{c}\sum_{i=0}^{k}\sum_{j=0}^{k}{g(i, j, m) \times h(x-i, y-j, m)},
\end{equation}
where $g$ and $h$ represent the kernel and feature map respectively, $f(x,y)$ is the calculated output of the 2D convolution in spatial coordinate $(x,y)$, $k$ is the kernel size, and $c$ is the channel number.
Note the common 2D operation regards different channels as same importance when aggregating them.
However, we argue that, when temporal information is squeezed into channels, it is necessary to distinguish their temporal importance.
The improved \textbf{T}emporal \textbf{F}ocus 2D \textbf{C}onvolution (TFC) is formulated by:
\begin{equation}
	\setlength{\abovedisplayskip}{-3pt}
	\setlength{\belowdisplayskip}{2pt}
	f^{\prime}(x,y)=\sum_{m=0}^{c}\sum_{i=0}^{k}\sum_{j=0}^{k}w_{m}\times{g(i, j, m) \times h(x-i, y-j, m)},
\end{equation}
where $w_m$ is the temporal-adaptive weights calculated according to the input features, it models the temporal importance of different channels.
$w_m$ can be computed using a lightweight module, \ie, weight computation module.
In this paper, we simply use a global MaxPool2d followed by a two-layer MLP as the WCM.
\par
\noindent\textbf{Inter-temporal Object Interaction Module.} The IOI module is designed from two aspects of consideration.
(1) On the one hand, when the temporal information of a video clip is squeezed into channels, and after being processed by a stem layer, the temporal order information of channels is mixed up and some discriminate details are lost.
The model needs to restore such temporal details.
(2) On the other hand, we argue that it is important to capture the relation among multiple objects attached to different temporal channels. Because, for the squeezed modality, two objects that interact with each other over time may appear in different channels.
The model also needs to learn such spatial relations.
IOI module is designed to restore such temporal positional dynamics and temporal object relations (right part of Formula \ref{equ:restore}):
\begin{equation}
	\setlength{\abovedisplayskip}{3pt}
	\setlength{\belowdisplayskip}{3pt}
	\xi=\phi\left(\eta^{T\rightarrow C_o}\left( \varphi\left( \eta^{C_i\rightarrow T}\left( \mathbf{F}_b \right)+\mathbf{F_T} \right)  \right) \right)\otimes\eta^{C_i\rightarrow C_o}\left(\mathbf{F}_b \right),
\end{equation}
where $\phi$ is a sigmoid function, $\eta$ is a mapping function that converts the number of channels, $T$ is the number of frames, $\eta$ is a function that excavates the inter-temporal object representations, $\mathbf{F_T}$ is the temporal position encoding, and $\otimes$ is the elementwise multiplication.  
\par
In this paper, we implement the above formula based on pure 2D convolution.
As shown in \figref{fig:pipe} (c), the IOI module consists of two sub-branches. 
The top branch first uses a $3\times 3$ TFC to reduce the number of channels ($C$) to the number of frames ($T$) and to capture the temporal importance, simultaneously.
Then temporal position encoding information is injected to restore the temporal dynamics.   
After this, a $7\times 7$ convolution is leveraged to model the object relations between $T$ frames.
Note it can be replaced by other more efficient modules to capture the cross-temporal object interactions if possible in the future.
The convolution with large kernels, in this work, is the simplest and also an effective way to model the cross-temporal object interactions.
Finally, a $3\times 3$ convolution is used to get the output number of channels.
The bottom branch makes a direct mapping from input channels to output channels.

\vspace{-15pt}
\section{Experiments}
\vspace{-10pt}
\subsection{Datasets}
\vspace{-10pt}
We evaluate the proposed method on $6$ datasets in total.
For action recognition, we conduct experiments on $4$ commonly-used datasets including Kinetics400 (K400)~\cite{k400}, Kinetics600 (K600)~\cite{k600}, and HMDB51~\cite{hmdb51}.
K400 is a large-scale action recognition benchmark with $400$ action classes, it contains $\sim 240k$ training videos and $\sim 20k$ validation videos.
K600 is an extension of the K400 dataset, which includes $600$ classes (each class with at least $600$ clips).
HMDB51 includes $6,849$ clips divided into $51$ action categories, each with a minimum of $101$ clips.
For action detection, we experiment on AVA2.1~\cite{ava} dataset.
AVA2.1 densely annotates $80$ atomic actions by localizing them in space and time in $430$ 15-minute movie clips.
It generates $1.62$M action labels with per human multiple labels occurring frequently.
Besides, we conduct experiments on THUMOS14~\cite{thumos14}, which contains $200$ and $212$ untrimmed videos for training and testing.

\vspace{-10pt}
\subsection{Implementation Details}
\vspace{-10pt}
In this work, we train and test the model using PyTorch framework on 8 NVIDIA Tesla V100 GPUs.
Following the configurations of ~\cite{resnet}, we set the numbers of CTL blocks of the proposed~\ourmodel~in four stages to [3, 4, 6, 3], and set the channel numbers of four stages to [256, 512, 1024, 2048].
The input frames of the proposed model are set to 16 for all experiments.
For K400~\cite{k400} and K600~\cite{k600} datasets, we train the mode using SGD optimizer with CosineAnnealing learning rate.
The initial learning rate is $0.015$, the warm-up epoch number is set to $8$, and weight decay is set to $7$e-5.
The model is initialized by pre-trained models on ImageNet1K~\cite{imagenet}.
The total epoch number is $100$ and the batch size is $512$.
The resolution of the input frame is $224\times 224$.
Following the common practice~\cite{tada}, we use a dense sampling strategy with the frame interval of $4$.
Following ~\cite{mvitv2,tada,videoswin}, we use multi-scale cropping, random flip, color jitter, and random erasing to augment the training data.
For testing, we use three spatial crops with $10$ temporal clips ($3\times 10$) strategy to evaluate the model, and the shorter side of each video is resized to $224$.
\par
For the HMDB51~\cite{hmdb51} dataset, we use a multi-step learning rate strategy with total epochs of $50$, an initial learning rate of $0.015$, and milestones of [12, 24].
The weight decay is set to $1$e-$2$.
We use $3$ crops $2$ clips strategy to evaluate the model.
Others are the same as K400.
For the AVA2.1~\cite{ava} dataset, we use SGD optimizer and set the total epoch number, initial learning rate, and weight decay to $50$, $0.01$, and $1$e-$5$ respectively.
Other settings are same as ~\cite{slowfast}.
For the THUMOS14~\cite{thumos14}, we use the default training settings as AFSD~\cite{afsd}.
To compare with other methods in fair, we use the released codes of their paper or mmaction2~\cite{mmaction2} framework to conduct experiments. 
\vspace{-10pt}
\subsection{Ablation Study}
\vspace{-5pt}
\textbf{Ablation analysis of two branches of CTL Module.}
To demonstrate the effectiveness of the two branches of the CTL module, we conduct ablation experiments on K400~\cite{k400}.
The base model is a simple model that replaces the CTL into a $3\times3$ convolution.
As shown in \tabref{tab:ablation_CTL}, the TFConv branch (branch1) achieves $62.8\%$ Top1 accuracy.
The accuracy of the IOI branch (branch2) is $69.6\%$, which is $6.8\%$ higher than the TFConv branch.
It is because TFConv branch only concentrates on extracting temporal importance of multiple channels using $1\times 1$ convolution while the IOI branch captures both the temporal importance and spatial relation using large kernels.
As shown in the last row, when combining the two branches, the performance of the proposed~\ourmodel~can reach $71.6\%$, which is $1.0\%$ higher than only using the IOI branch and $4.4\%$ higher than the base model.

\begin{table}%
	\small
	\vspace{-5pt}
	\begin{minipage}[t]{0.5\textwidth}
		\centering
		\makeatletter\def\@captype{table}
		\caption{Ablation of two branches of CTL Module on K400~\cite{k400}.}
		\vspace{-5pt}
		\setlength{\tabcolsep}{0.6mm}
		\renewcommand\arraystretch{0.9}
		\begin{tabular}{l|cccc}
			\toprule
			
			\textbf{Settings} &\textbf{Param}& \textbf{FLOPs} & \textbf{Top1} & \textbf{Top5}  \\
			\midrule
			base&23.6M&4.7G&67.2&87.1\\
			TFConv Branch &16.1M& 3.1G&62.8&83.5\\
			IOI Branch &27.5M&5.3G&69.6&88.3\\
			Ensemble (\ourmodel)&28.7M&5.5G&\textbf{71.6}&\textbf{89.3}\\
			\bottomrule
			
		\end{tabular}%
		\label{tab:ablation_CTL}%
	\end{minipage}
	\begin{minipage}[t]{0.5\textwidth}
		\centering
		\makeatletter\def\@captype{table}
		\caption{Ablation analysis of IOI branch of CTL Module on K400~\cite{k400} dataset.%
		}
		\vspace{-5pt}
		\setlength{\tabcolsep}{0.6mm}
		\renewcommand\arraystretch{0.9}
		\begin{tabular}{l|ccc|cccc}
			\toprule
			\textbf{\#}&IOI&TFC&TPosEmb &\textbf{Param}& \textbf{FLOPs} & \textbf{Top1} & \textbf{Top5}  \\
			\midrule
			1&$\times$&$\times$&$\times$&23.6M&4.7G&67.2&87.1\\
			2&$\checkmark$&$\times$&$\times$&24.9M&5.3G&68.1&87.4\\
			3&$\checkmark$&$\checkmark$&$\times$&27.4M&5.3G&69.2&88.4\\
			4&$\checkmark$&$\checkmark$&$\checkmark$&27.5M&5.3G&69.6&88.3\\
			\bottomrule
			
		\end{tabular}%
		\label{tab:ablation_ioi}%
	\end{minipage}
\vspace{-5pt}
\end{table}
\begin{table}%
	\small
	\begin{minipage}[t]{0.5\textwidth}
		\centering
		\makeatletter\def\@captype{table}
		\caption{Ablation analysis of temporal focus convolution (TFC) of CTL Module on K400~\cite{k400}.}
		\vspace{-5pt}
		\setlength{\tabcolsep}{1.0mm}
		\renewcommand\arraystretch{0.9}
		\begin{tabular}{l|cccc}
			\toprule
			\textbf{Settings} &\textbf{Param}& \textbf{FLOPs} & \textbf{Top1} & \textbf{Top5}  \\
			\midrule
			$1\times 1$ Convolution&13.5M& 3.1G&49.9&73.3\\
			Temporal focus conv&16.1M&3.1G&62.8&83.5\\
			\bottomrule
			
		\end{tabular}%
		\label{tab:ablation_taconv}%
	\end{minipage}
	\begin{minipage}[t]{0.5\textwidth}
		\centering
		\makeatletter\def\@captype{table}
		\caption{Effect of the input frames of~\ourmodel.
		}
		\vspace{-5pt}
		\setlength{\tabcolsep}{1.6mm}
		\renewcommand\arraystretch{0.93}
		\begin{tabular}{l|cccc}
			\toprule
			\textbf{Frame Number} &\textbf{4}& \textbf{8} & \textbf{16} & \textbf{32}  \\
			\midrule
			FLOPs (G)&4.8&5.0&5.5&6.5\\
			Top1 Acc ($\%$)&70.7&70.9&71.6&71.7 \\
			Top5 Acc ($\%$)&89.1&88.9&89.3&89.4 \\
			\bottomrule
			
		\end{tabular}%
		\label{tab:ablation_frame}%
	\end{minipage}
\vspace{-15pt}
\end{table}

\noindent\textbf{Effectiveness analysis of the temporal focus convolution (TFC).} The TFConv branch of the CTL module is to model the temporal importance of multiple channels.
In this experiment, we analyze by only using the TFConv branch of the CTL Module.
As shown in \tabref{tab:ablation_taconv}, when not using the temporal focus convolution (TFC), \ie, only using $1\times 1$ convolution, the accuracy is only $49.9\%$.
When using the temporal focus convolution, the accuracy is increased by $12.9\%$.
It demonstrates that temporal focus convolution is important when squeezing the temporal axis into channels.

\noindent\textbf{Ablation analysis of the IOI branch of CTL Module.} We make an ablation analysis of the IOI branch in \tabref{tab:ablation_ioi}.
In this experiment, we make an ablation on the model only using the IOI branch (\ie, the third row of \tabref{tab:ablation_CTL}). 
As shown in \tabref{tab:ablation_ioi}, the base model in the first row is only a $3\times 3$ convolution (see the bottom part of \figref{fig:pipe} (c)) that simply maps the input channels into output channels.
We then gradually add IOI, temporal focus convolution, and temporal position encoding into the base model. 
First, the performance of the base model is $67.2\%$.
After adding the top branch of IOI, \ie, the second row of \tabref{tab:ablation_ioi}, the Top1 accuracy is increased to $68.1\%$ ($+0.9\%$), which demonstrates the cross-temporal object relation is important for the model.
Then after replacing the first $3\times 3$ convolution of the top part into a $3\times 3$ TFC (\ie, the third row), the Top1 accuracy is further increased to $69.2\%$ ($+0.9\%$).
Finally, the temporal position encoding further improves the performance by $0.4\%$, which indicates that temporal order information is also necessary for the model to capture temporal representations.

\noindent\textbf{Impact of the number of input frames.} Here we investigate the effects of changing the number of input frames.
When changing the different number of input frames, the input channel number of the $5\times 5$ convolution of the stem layer (see the top part of \figref{fig:pipe}) is changed accordingly. The sampling interval of frames $4$, $8$, $16$ and $32$ are $12$, $8$, $4$ and $2$, respectively.
Other models and training settings are unchanged.
As shown in \tabref{tab:ablation_frame}, using $4$ and $8$ frames, the Top1 accuracies are $70.7\%$ and $70.9\%$, respectively.
When using $16$ frames, the performance is increased to $71.6\%$ ($+0.7\%$).
The $32$ frames have the best $71.7\%$ Top1 accuracy.
Considering the data loading time consuming of $32$ frames is much more than $16$ frames, we finally use $16$ input frames in the proposed~\ourmodel.

\begin{table}%
	\vspace{-15pt}
	\small
	\begin{minipage}[t]{0.35\textwidth}
		\makeatletter\def\@captype{table}
		\caption{Impact of different channel expansion of~\ourmodel. 
		}
		\vspace{-5pt}
		\setlength{\tabcolsep}{0.2mm}
		\renewcommand\arraystretch{0.9}
		\begin{tabular}{l|cccc}
			\toprule
			\textbf{Channel factor} &\textbf{0.5}& \textbf{0.75} & \textbf{1.0} & \textbf{1.25}   \\
			\midrule
			FLOPs (G)&1.9&3.4&5.5&8.1\\
			Tput (clips / s)&1964&1186&903&617\\
			Top1 Acc ($\%$)&67.8&69.8&71.6&71.9 \\
			\bottomrule
			
		\end{tabular}%
		\label{tab:ablation_channels}%
	\end{minipage}
	\begin{minipage}[t]{0.65\textwidth}
		\centering
		\makeatletter\def\@captype{table}
		\caption{Latency of models on CPU of a modern mobile phone. Top1 is on K400 dataset. `LAT' is latency.
		}
		\vspace{-5pt}
		\setlength{\tabcolsep}{0.1mm}
		\renewcommand\arraystretch{0.9}
		\begin{tabular}{l|ccccccc}
			\toprule
			Model    &I3d &MoViNet &TSN &E3D &TSM &TimeSFormer &SqueezeTime \\
			Backbone&Res18&A0&Res34&XS&Res34&ViT-B&-\\
			\midrule
			Top1(\%) &{65.3} &{65.8} &{67.7} &{68.6} &{69.4} &{70.4} &\textbf{71.6} \\
			LAT(ms)  &869 &1282 &1239 &819 &688 &960 &\textbf{139} \\
			\bottomrule
		\end{tabular}%
		\label{tab:mobile_test}%
	\end{minipage}
	\vspace{-15pt}
\end{table}
\begin{table*}[t]
	\footnotesize
	\caption{Performance comparison of multiple lightweight methods on K400~\cite{k400} dataset. CPU latency (LAT, ms), GPU throughput (Tput, clips / s), Top1 accuracy, and Top5 accuracy are reported. We retrain and test these methods using mobile settings, \ie, low FLOPs, low CPU Latency, and high GPU throughput, by adopting a smaller backbone, lower clip resolution, or fewer frames to build comparable models.
		The latency and throughput are tested on an Intel(R) Xeon(R) Gold 6278C@2.60GHz CPU (batch size of 1) and an NVIDIA Tesla V100 GPU (using maximum batch size). {{The above settings are the same for the following several tables.}}
	}
	
	\setlength{\tabcolsep}{0.4mm}
	\renewcommand\arraystretch{0.8}
	\begin{tabular}{lccccccc}
		\toprule
		{\textbf{Model}}                  & \textbf{Param(M)} & \textbf{FLOPs(G)}& \textbf{Frames}& \textbf{LAT(CPU)}$\downarrow$  & \textbf{Tput(GPU)}$\uparrow$ & \textbf{Top1(\%)}   & \textbf{Top5(\%)} \\
		\midrule
		I3d-R18~\cite{i3d}                           & 33.4             & 46.0                    & 32$\times$3$\times$10                      & 450                                 & 128              & 65.3                                 & 86.6             \\
		MoViNet-A0~\cite{movinets}                      & 3.1              & 2.7                    & 50$\times$1$\times$1                       & 505                                 & 108              & 65.8                                 & 87.4             \\
		TSM-R18~\cite{tsm}                              & 11.2             & 19.0                  & 8$\times$8$\times$10                       & 186                                 & 429              & 65.9                                 & 86.6             \\
		X3d-XS~\cite{x3d}                                     & 3.80             & 0.9                     & 4$\times$3$\times$10                       & 219                                 & 441              & 66.6                                 & 86.7                 \\
		TADA-R18~\cite{tada}                            & 14.2             & 19.2                    & 8$\times$3$\times$10                       & 767                                 & 76               & 66.9                                 & 87.5             \\
		TSN-R34~\cite{tsn}                            & 21.5             & 91.8                    & 25$\times$25$\times$10                     & 636                                 & 108              & 67.7                                 & 88.0             \\
		R2plus1d-R18~\cite{r2plus1d}                       & 33.5            & 30.4                   & 8$\times$3$\times$10                       & 510                                 & 133              & 68.1                                & 87.7  \\
		C2d-R34~\cite{nonlocal}                        & 21.5             & 17.8                    & 8$\times$3$\times$10                       & 205                                 & 455              & 68.5                                 & 88.0           \\
		{SlowOnly-R18}~\cite{slowfast}                & 32.0             & 35.9                    & 8$\times$3$\times$10                       & 303                                 & 164              & 68.6                                 & 88.3             \\
		E3d-XS~\cite{e3d}                                      & 2.2              & 2.4                     & 16$\times$3$\times$10                      & 313                                 & 203              & 68.6                                 & 88.3             \\
		Slowfas-R18~\cite{slowfast}                      & 33.6             & 25.2                    & 32$\times$3$\times$10                      & 604                                 & 116              & 69.0                                 & 88.9             \\
		TIN-R34~\cite{tin}                              & 21.5             & 29.4                    & 8$\times$8$\times$1                        & 345                                 & 274              & 69.1                                 & 88.5             \\
		TSM-R34~\cite{tsm}                            & 21.5             & 29.4                   & 8$\times$8$\times$10                       & 238                                 & 307              & 69.4                                 & 88.8             \\
		TANet-R18~\cite{tam}                           & 11.8             & 19.1                   & 8$\times$3$\times$8                        & 192                                 & 379              & 69.6                                 & 88.7             \\
		R2plus1d-R34~\cite{r2plus1d}                      & 63.8             & 53.1                    & 8$\times$3$\times$10                       & 748                                 & 82               & 69.8                                 & 88.4             \\
		TPN-R18~\cite{tpn}                             & 38.1             & 36.7                    & 8$\times$3$\times$10                       & 344                                 & 158              & 70.2                                 & 89.3             \\
		\midrule
		MViTv2-Small~\cite{mvitv2}                              & 34.5             & 4.2                   & 4$\times$3$\times$10                       & 176                                 & 517              & 68.8                                 & 88.2             \\
		VideoSwin-Tiny~\cite{videoswin}                     & 28.2             & 5.6                    & 4$\times$3$\times$10                       & 162                                 & 505              & 69.9                                 & 88.9             \\
		TimeSformer-Base~\cite{timeformer}                    & 86.0             & 23.2                   & 4$\times$3$\times$10                       & 196                                 & 205              & 70.4                                 & 89.2             \\
		\midrule
		\rowcolor{blue!15}SqueezeTime (Ours)                                    & 28.7            & 5.5                    & 16$\times$3$\times$10                      & { \textbf{123}}                     & \textbf{903}     &\textbf{71.6}                         &\textbf{89.3}            \\
		\bottomrule    
	\end{tabular}
	\label{tab:res_k400}
	\vspace{-20pt}
\end{table*}
\noindent\textbf{Impact of the expansion of channels of~\ourmodel}. Note we squeeze the temporal axis of a video sequence into the spatial channel dimension, thus we research the effect of channel numbers of the proposed~\ourmodel~in \tabref{tab:ablation_channels}.
The base channel factor $1.0$ represents that the output channels of four stages are [$256, 512, 1024, 2048$].
Other factors are changed according to the factor.
The number of channels are changed according to the expansion factor. 
As shown in the table, the expansion factor $1.0$ achieves the best balance between the accuracy and GPU throughput.
\par
\noindent\textbf{Speed evaluation of different methods on mobile devices.} As shown in \tabref{tab:mobile_test}, we provided the latencies of multiple methods on the mobile device. As shown in the table, the proposed SqueezeTime is much more efficient for the constrained hardware, \eg, {{six times faster} (only 139ms) than TimesFormer (960ms), on the mobile device.}

\begin{table*}[t]
	\centering
	\footnotesize
	\caption{Performace comparison of lightweight methods on K600~\cite{k600} dataset.
		LAT: CPU Latency (ms), Tput: GPU Throughput (clips/s).
	}
	\setlength{\tabcolsep}{0.3mm}
	\renewcommand\arraystretch{0.8}
	\begin{tabular}{lccccccc}
		\toprule
		\multicolumn{1}{l}{\textbf{Model}}  & \textbf{Param(M)} & \textbf{FLOPs(G)} & \textbf{Frames} & \textbf{LAT(CPU)}$\downarrow$ & \textbf{Tput(GPU)}$\uparrow$ & \textbf{Top1(\%)} & \textbf{Top5(\%)} \\
		\midrule
		I3d-R18~\cite{i3d}              & 33.4  & 46.0            & 32$\times$3$\times$10     & 450   & 128  & 70.7  & 90.4  \\
		TSN-R34~\cite{tsn}               & 21.5  & 91.8            & 25$\times$25$\times$10     & 636   & 108  & 71.4  & 90.2  \\
		
		MoViNet-A0~\cite{movinets}           & 3.1   & 2.7          & 50$\times$1$\times$1     & 505   & 108  & 71.5  & 90.4  \\
		X3d-XS~\cite{x3d}                            & 3.8   & 0.9              & 4$\times$3$\times$10      & 219   & 441  & 71.5  & 90.6  \\
		TIN-R34~\cite{tin}             & 21.5  & 29.4            & 8$\times$8$\times$1      & 345   & 274  & 71.6  & 90.7  \\
		C2d-R34~\cite{nonlocal}          & 21.5  & 17.8           & 8$\times$3$\times$10      & 205   & 455  & 72.1  & 90.8  \\
		
		SlowOnly-R18~\cite{slowfast}  & 32.0  & 35.9    & 8$\times$3$\times$10    & 303   & 164  & 72.2  & 91.1  \\
		E3d-XS~\cite{e3d}               & 2.2   & 2.4             & 16$\times$3$\times$10     & 313   & 203  & 72.9  & 91.5  \\  
		Slowfast-R18~\cite{slowfast}    & 33.6  & 25.2      & 32$\times$3$\times$10     & 604   & 116  & 73.5  & 91.9  \\
		
		TSM-R34~\cite{tsm}                 & 21.5  & 29.4            & 8$\times$8$\times$10      & 238   & 307 & 73.5  & 91.3  \\
		TANet-R18~\cite{tam}               & 11.8  & 19.1          & 8$\times$8$\times$10      & 192   & 379  & 73.8  & 91.9  \\
		TPN-R18~\cite{tpn}                 & 38.1  & 36.7            & 8$\times$3$\times$10      & 344   & 158  & 73.8  & 91.9  \\
		TADA-R18~\cite{tada}               & 14.2  & 19.2           & 8$\times$3$\times$10      & 767   & 77  & 73.9  & 92.0  \\
		R2plus1d-R18~\cite{r2plus1d}          & 33.5  & 30.4       & 8$\times$3$\times$10      & 510   & 133  & 74.0  & 92.0  \\
		R2plus1d-R34~\cite{r2plus1d}         & 63.8  & 53.1       & 8$\times$3$\times$10      & 748   & 82  & 75.5  & 92.6  \\ 
		\midrule
		TimeSformer-Base~\cite{timeformer}    & 86.0  & 23.2        & 4$\times$3$\times$10      & 196   & 205  & 73.5  & 91.4  \\
		VideoSwin-Tiny~\cite{videoswin}         & 28.2  & 5.6           & 4$\times$3$\times$10      & 162   & 505  & 73.6  & 91.4  \\
		MViTv2-Small~\cite{mvitv2}                & 34.5  & 4.2               & 4$\times$3$\times$10      & 176   & 517  & 73.6  & 91.7  \\
		\midrule
		\rowcolor{blue!15}SqueezeTime (Ours)  & 28.7  & 5.5    & 16$\times$3$\times$10     & \textbf{123} & \textbf{903}  &\textbf{76.0}  &92.5 \\
		\bottomrule
	\end{tabular}%
	\label{tab:res_k600}%
	\vspace{-20pt}
\end{table*}%
\vspace{-10pt}
\subsection{Comparisons with the State of the Arts}
\vspace{-5pt}
In this section, we conduct comprehensive experiments to compare our~\ourmodel~with state-of-the-art methods in mobile video analysis settings.
\par
\noindent\textbf{Comparison of action recognition on K400~\cite{k400}.} As shown in \tabref{tab:res_k400}, we compare the proposed~\ourmodel~with $19$ state-of-the-art methods.
Note we aim to build a lightweight and fast backbone, \ie, with low CPU latency, high GPU throughput, and reliable accuracy, for mobile video analysis, thus we retrain and test these compared models using mobile settings for a fair comparison.
We convert these models into mobile versions by adopting smaller backbones, using lower video resolution, or using fewer frames.
As shown in \tabref{tab:res_k400}, the top rows are CNN-based models and the middle rows are transformer-based models.
The best comprehensive CNN network and transformer network are TANet~\cite{tam} and VideoSwin~\cite{videoswin} respectively.
The proposed~\ourmodel~performs better than all compared models in terms of CPU latency, GPU Throughput, and Top1 accuracy.
For example, it exceeds the best compared transformer-based model VideoSwin by $1.7\%$ Top1 accuracy but with $\sim 25\%$ lower CPU latency and $\sim 80\%$ higher GPU throughput and exceeds the best compared CNN-based model TANet by $2.0\%$ with $\sim 25\%$ lower CPU latency and $\sim 140\%$ higher GPU throughput.
From the table, we can also find the 2D-based CNN models, \eg, TANet and TSM are more mobile friendly (faster) than the 3D-based CNN models, \eg, I3d, X3d, MoViNet, and E3d.
Although the transformer-based models show competitive performance compared with these 2D-based CNN models, the proposed~\ourmodel~using 2D convolution can also beat them in mobile settings, which demonstrates the effectiveness of our model. 
\par
\noindent\textbf{Comparison of action recognition on K600~\cite{k600}.} As shown in \tabref{tab:res_k600}, We compare the proposed~\ourmodel~with state-of-the-art models on K600.
As shown in the table, our model significantly outperforms other methods with at least $0.5\%$ Top1 accuracy improvement, $\sim 25\%$ CPU speed improvement, and $\sim 80\%$ GPU speed improvement.
Note that although R2plus1d with ResNet34 backbone achieves the best Top1 accuracy, \ie, $75.5\%$, among all compared methods on K600, its CPU speed, and GPU speed are only $\sim 1/6$ and $\sim 1/9$ of our proposed~\ourmodel.
\par
\begin{table*}[t]
	\centering
	\footnotesize
	\caption{Performace comparison of action recognition on HMDB51~\cite{hmdb51} dataset (Pretrained on K400). LAT: CPU Latency (ms), Tput: GPU Throughput (clips/s).}
	\setlength{\tabcolsep}{0.2mm}
	\renewcommand\arraystretch{0.8}
	\begin{tabular}{lccccccc}
		\toprule
		\multicolumn{1}{l}{\textbf{Model}} & \textbf{Param(M)} & \textbf{FLOPs(G)} & \textbf{Frames}$\times$ \textbf{Views} & \textbf{LAT(CPU)}$\downarrow$ & \textbf{Tput(GPU)}$\uparrow$ & \textbf{Top1(\%)} & \textbf{Top5(\%)} \\
		\midrule
		X3d-XS~\cite{x3d}           & 3.8   & 0.9        & 4$\times$3$\times$2     & 219   & 441   & 57.1  & 85.4  \\
		E3d-XS~\cite{e3d}        & 2.2   & 1.2      & 4$\times$3$\times$2     & 211   & 415   & 57.7  & 86.3  \\
		SlowOnly-R18~\cite{slowfast}    & 32.0  & 18.0     & 8$\times$3$\times$2     & 268   & 321   & 61.2  & 88.5  \\
		I3d-R18~\cite{i3d}        & 33.4  & 46.0     & 32$\times$3$\times$2    & 450   & 128   & 61.3  & 87.3  \\
		TSN-R34~\cite{tsn}         & 21.5  & 91.8     & 8$\times$3$\times$2   & 636   & 108   & 61.4  & 89.1  \\
		TSM-R18~\cite{tsm}        & 11.2  & 19.0     & 8$\times$3$\times$2     & 186   & 429   & 61.6  & 88.4  \\
		R2plus1d-R34~\cite{r2plus1d}    & 63.8   &53.1    &8$\times$3$\times$2 &748&82&62.0&86.8\\
		TIN-R34~\cite{tin}         & 21.5  & 29.4     & 8$\times$3$\times$2     & 345   & 274   & 63.0  & 87.7  \\
		TANet-R18~\cite{tam}       & 11.8  & 19.1     & 8$\times$3$\times$2     & 192   & 379   & 63.8  & 89.8  \\
		C2d-R34~\cite{nonlocal}        & 21.5  & 17.8     & 8$\times$3$\times$2     & 205   & 455   & 63.9  & 90.9  \\
		Slow-fast-R18~\cite{slowfast}   & 33.8  & 19.3   & 16$\times$3$\times$2    & 473   & 157   & 64.1  & 89.0  \\
		E3d-XS~\cite{e3d}         & 2.2   & 2.4      & 16$\times$3$\times$2    & 313   & 203   & 65.0  & 91.1  \\
		TPN-R18~\cite{tpn}         & 37.4  & 28.1    & 8$\times$3$\times$2     & 256   & 203   & 65.2  & 89.4  \\
		\midrule
		VideoSwin-Tiny~\cite{videoswin}        & 28.2  & 5.6      & 4$\times$3$\times$2     & 162   & 505   & 61.1  & 86.4  \\
		MViTv2-Small~\cite{mvitv2}          & 34.5  & 4.2     & 4$\times$3$\times$2     & 176   & 517   & 62.2  & 88.2  \\
		Timesformer-Base~\cite{timeformer}      & 86.0  & 23.2    & 4$\times$3$\times$2     & 196   & 205   & 63.5  & 89.8  \\
		\midrule
		\rowcolor{blue!15}SqueezeTime (Ours)   & 28.7  & 5.5    & 16$\times$3$\times$2    & \textbf{123} & \textbf{903}   & \textbf{65.6} & 90.3  \\
		\bottomrule
	\end{tabular}%
	\label{tab:res_hmdb51}%
	\vspace{-5pt}
\end{table*}%

\begin{table}%
	\vspace{-0pt}
	\begin{minipage}[t]{0.45\textwidth}
		\centering
		\makeatletter\def\@captype{table}\makeatother\caption{Action detection results on AVA2.1~\cite{ava}. We train and test these methods using the same settings as ~\cite{slowfast}.
	}
		\vspace{-5pt}
		\setlength{\tabcolsep}{0.3mm}
		\renewcommand\arraystretch{0.8}
		\begin{tabular}{lccc}
			\toprule
			\multicolumn{1}{c}{\textbf{Model}} & \textbf{Frame} & \textbf{mAP}&\textbf{Time} \\
			\midrule
			SlowOnly-R18~\cite{slowfast}   & 8  & 10.8&3.9 ms  \\
			X3d-XS~\cite{x3d}   & 16  & 13.6&9.8 ms  \\
			E3d-XS~\cite{e3d}    & 16 & 14.8&7.3 ms  \\
			VideoSwin~\cite{videoswin}   & 16  & 12.7&8.6 ms  \\
			I3d-R18~\cite{i3d}   & 16  & 15.2&6.3 ms  \\
			R2plus1d-R34~\cite{r2plus1d}     & 16  &15.4&11.2 ms  \\
			\midrule
			SqueezeTime (Ours)   & 16  & 15.1&\textbf{3.4} ms  \\
			\bottomrule
		\end{tabular}%
	\label{tab:res_ava}%
	\end{minipage}
	\begin{minipage}[t]{0.55\textwidth}
		\centering
		\makeatletter\def\@captype{table}\makeatother\caption{Temporal action localization results of models on THUMOS14~\cite{thumos14} dataset.
			We measure the performance by mAP at different tIOU thresholds and average mAP in [0.3:0.1:0.7].
			We use the framework of AFSD~\cite{afsd} to train and test these methods.
			`Time' is the average time of predicting a test sample.}
		\vspace{-5pt}
		\setlength{\tabcolsep}{0.3mm}
		\renewcommand\arraystretch{0.8}
		\begin{tabular}{lccccccc}
			\toprule
			\textbf{Model} & \textbf{0.3} & \textbf{0.4} & \textbf{0.5} & \textbf{0.6} & \textbf{0.7} & \textbf{Avg} &\textbf{Time} \\
			\midrule
			I3d-R18~\cite{i3d} & 42.7  & 37.6  & 31.4  & 23.2  & 14.8  & 29.9&10.9s \\
			X3d-XS~\cite{x3d} &45.6    &40.4   &32.5   &22.6   &13.4  &30.9&2.9s\\
			E3d-XS~\cite{e3d}   &46.6     &41.6   &33.7   &24.8   &15.3  &32.4&1.4s \\
			\rowcolor{blue!15}SqueezeTime (Ours)  & 48.3  & 43.0  & 34.8  & 24.7  & 13.8  & \textbf{32.7}&\textbf{1.2}s \\
			\bottomrule
		\end{tabular}%
			\label{tab:res_thumos14}%
	\end{minipage}
\vspace{-15pt}
\end{table}

\noindent\textbf{Comparison of action recognition on HMDB51~\cite{hmdb51}.} As shown in \tabref{tab:res_hmdb51}, we compare our model with $16$ other methods on HMDB51.
From the table, we can find that TPN reaches the best Top1 accuracy ($0.2\%$ better than ours), but its CPU speed and GPU speed are only approximate $1/2$ and $1/4$ of our model.
Our proposed~\ourmodel~also shows competitive performance on this dataset.
\par
\noindent\textbf{Comparison of action detection on AVA2.1~\cite{ava}}. 
In this experiment, we demonstrate the effectiveness of~\ourmodel~on downstream AVA action detection task in \tabref{tab:res_ava}.
For the AVA2.1 dataset, spatial-temporal labels are provided for one frame per second, and each person was annotated with a bounding box and his actions.
We follow the standard protocol of~\cite{slowfast} to train the model and evaluate the performance on $60$ chasses.
We report the mean average precision (mAP) over $60$ classes, using a frame-level IoU threshold of $0.5$.
We experiment using mmaction2~\cite{mmaction2} framework and replace the backbone of~\cite{slowfast} using the proposed~\ourmodel.
To align the deep features of the backbone, we simply reshape the output of~\ourmodel, \ie, ($C_{out},\mathbf{F}_h,\mathbf{F}_w$), to ($C_{out}//16,16,\mathbf{F}_h,\mathbf{F}_w$), where $C_{out}$, $\mathbf{F}_h$ and $\mathbf{F}_w$ are the channel number, height and width of the feature map.
The number $16$ is the temporal dimension of the reshaped feature.
The proposed~\ourmodel~shows competitive performance in \tabref{tab:res_ava}.
Although the mAP of our model is $0.3$ lower than the best compare R2plus1d-R34~\cite{r2plus1d}, the average test time on the test set of our model is only $\sim 1/3$ of it, which is much more efficient for the mobile usage.
\par
\noindent\textbf{Comparison of action detection on THUMOS14~\cite{thumos14}.} In this part, we investigate the effectiveness of the proposed~\ourmodel~on the action detection dataset of THUMOS14~\cite{thumos14}.
In this experiment, we compare the proposed model with $3$ popular models based on the framework of \cite{afsd}. We replace the original backbone of \cite{afsd} using these models and remove the boundary refinement mechanism, \ie, the models finally contain only a backbone and a prediction head.
We adjust the number of input frames, and resolution of videos of these models to make fair comparisons in mobile settings.
The other training and test settings are as default as \cite{afsd}.
As for the proposed~\ourmodel, we make some designs to make it suitable for action detection tasks with a long temporal sequence.
For the $256$ input frames, we split them along the temporal dimension using a temporal sliding window of $16$ frames and the stride is $8$.
Then we get $32$ temporal clips with the frame number of $16$ and we feed them to~\ourmodel~in parallel to get features of a temporal dimension of $32$.
Then the features are fed into the detection head to predict the result.
As shown in \tabref{tab:res_thumos14}, the proposed~\ourmodel~can exceed all $3$ compared methods in terms of the average mAP and test time.
It is $0.3$ better than the recently proposed E3d-XS but using $\sim 20\%$ fewer test time, which demonstrates the effectiveness of~\ourmodel~as a backbone for the downstream action detection task.

\vspace{-15pt}
\section{Conclusion}
\vspace{-10pt}
In this paper, we concentrate on building a lightweight and fast model for mobile video analysis.
Different from current popular video models that regard time as an extra dimension, we propose to squeeze the temporal axis of a video sequence into the spatial channel dimension, which saves a great amount of memory and computation consumption.
To remedy the performance drop caused by the squeeze operation, we elaborately design an efficient backbone~\ourmodel~with a stack of efficient Channel-Time Learning Block (CTL), which consists of two complementary branches to restore and excavate temporal dynamics.
Besides, we make comprehensive experiments to compare a quantity of state-of-the-art methods in mobile settings, which shows the superiority of the proposed~\ourmodel, and we hope it can foster further research on mobile video analysis.
{
	\small
	\bibliographystyle{unsrt}
	\bibliography{main}

\begin{thebibliography}{10}

\bibitem{resnet}
Kaiming He, Xiangyu Zhang, Shaoqing Ren, and Jian Sun.
\newblock Deep residual learning for image recognition.
\newblock In {\em CVPR}, pages 770--778, 2016.

\bibitem{x3d}
Christoph Feichtenhofer.
\newblock X3d: Expanding architectures for efficient video recognition.
\newblock In {\em CVPR}, pages 203--213, 2020.

\bibitem{k400}
Will Kay, Joao Carreira, Karen Simonyan, Brian Zhang, Chloe Hillier, Sudheendra
  Vijayanarasimhan, Fabio Viola, Tim Green, Trevor Back, Paul Natsev, et~al.
\newblock The kinetics human action video dataset.
\newblock {\em arXiv preprint arXiv:1705.06950}, 2017.

\bibitem{k600}
Joao Carreira, Eric Noland, Andras Banki-Horvath, Chloe Hillier, and Andrew
  Zisserman.
\newblock A short note about kinetics-600.
\newblock {\em arXiv preprint arXiv:1808.01340}, 2018.

\bibitem{hmdb51}
Hildegard Kuehne, Hueihan Jhuang, Estibaliz Garrote, Tomaso Poggio, and Thomas
  Serre.
\newblock Hmdb: a large video database for human motion recognition.
\newblock In {\em ICCV}, pages 2556--2563, 2011.

\bibitem{c3d}
Du~Tran, Lubomir Bourdev, Rob Fergus, Lorenzo Torresani, and Manohar Paluri.
\newblock Learning spatiotemporal features with 3d convolutional networks.
\newblock In {\em ICCV}, pages 4489--4497, 2015.

\bibitem{i3d}
Joao Carreira and Andrew Zisserman.
\newblock Quo vadis, action recognition? a new model and the kinetics dataset.
\newblock In {\em CVPR}, pages 6299--6308, 2017.

\bibitem{slowfast}
Christoph Feichtenhofer, Haoqi Fan, Jitendra Malik, and Kaiming He.
\newblock Slowfast networks for video recognition.
\newblock In {\em ICCV}, pages 6202--6211, 2019.

\bibitem{res3d}
Kensho Hara, Hirokatsu Kataoka, and Yutaka Satoh.
\newblock Can spatiotemporal 3d cnns retrace the history of 2d cnns and
  imagenet?
\newblock In {\em CVPR}, pages 6546--6555, 2018.

\bibitem{xie2018rethinking}
Saining Xie, Chen Sun, Jonathan Huang, Zhuowen Tu, and Kevin Murphy.
\newblock Rethinking spatiotemporal feature learning: Speed-accuracy trade-offs
  in video classification.
\newblock In {\em ECCV}, pages 305--321, 2018.

\bibitem{r2plus1d}
Du~Tran, Heng Wang, Lorenzo Torresani, Jamie Ray, Yann LeCun, and Manohar
  Paluri.
\newblock A closer look at spatiotemporal convolutions for action recognition.
\newblock In {\em CVPR}, pages 6450--6459, 2018.

\bibitem{movinets}
Dan Kondratyuk, Liangzhe Yuan, Yandong Li, Li~Zhang, Mingxing Tan, Matthew
  Brown, and Boqing Gong.
\newblock Movinets: Mobile video networks for efficient video recognition.
\newblock In {\em CVPR}, pages 16020--16030, 2021.

\bibitem{e3d}
Junyan Wang, Zhenhong Sun, Yichen Qian, Dong Gong, Xiuyu Sun, Ming Lin, Maurice
  Pagnucco, and Yang Song.
\newblock Maximizing spatio-temporal entropy of deep 3d cnns for efficient
  video recognition.
\newblock {\em ICLR}, 2023.

\bibitem{tsm}
Ji~Lin, Chuang Gan, and Song Han.
\newblock Tsm: Temporal shift module for efficient video understanding.
\newblock In {\em ICCV}, pages 7083--7093, 2019.

\bibitem{tdn}
Limin Wang, Zhan Tong, Bin Ji, and Gangshan Wu.
\newblock Tdn: Temporal difference networks for efficient action recognition.
\newblock In {\em CVPR}, pages 1895--1904, 2021.

\bibitem{tam}
Zhaoyang Liu, Limin Wang, Wayne Wu, Chen Qian, and Tong Lu.
\newblock Tam: Temporal adaptive module for video recognition.
\newblock In {\em ICCV}, pages 13708--13718, 2021.

\bibitem{adafocus}
Yulin Wang, Zhaoxi Chen, Haojun Jiang, Shiji Song, Yizeng Han, and Gao Huang.
\newblock Adaptive focus for efficient video recognition.
\newblock In {\em ICCV}, pages 16249--16258, 2021.

\bibitem{xiang2022spatiotemporal}
Wangmeng Xiang, Chao Li, Biao Wang, Xihan Wei, Xian-Sheng Hua, and Lei Zhang.
\newblock Spatiotemporal self-attention modeling with temporal patch shift for
  action recognition.
\newblock In {\em ECCV}, pages 627--644, 2022.

\bibitem{tada}
Ziyuan Huang, Shiwei Zhang, Liang Pan, Zhiwu Qing, Mingqian Tang, Ziwei Liu,
  and Marcelo~H Ang~Jr.
\newblock Tada! temporally-adaptive convolutions for video understanding.
\newblock In {\em ICLR}, 2022.

\bibitem{timeformer}
Gedas Bertasius, Heng Wang, and Lorenzo Torresani.
\newblock Is space-time attention all you need for video understanding?
\newblock page~4, 2021.

\bibitem{mvitv2}
Yanghao Li, Chao-Yuan Wu, Haoqi Fan, Karttikeya Mangalam, Bo~Xiong, Jitendra
  Malik, and Christoph Feichtenhofer.
\newblock Mvitv2: Improved multiscale vision transformers for classification
  and detection.
\newblock In {\em CVPR}, pages 4804--4814, 2022.

\bibitem{videoswin}
Ze~Liu, Jia Ning, Yue Cao, Yixuan Wei, Zheng Zhang, Stephen Lin, and Han Hu.
\newblock Video swin transformer.
\newblock In {\em CVPR}, pages 3202--3211, 2022.

\bibitem{aim}
Taojiannan Yang, Yi~Zhu, Yusheng Xie, Aston Zhang, Chen Chen, and Mu~Li.
\newblock Aim: Adapting image models for efficient video understanding.
\newblock In {\em ICLR}, 2023.

\bibitem{survey_on_pami}
Javier Selva, Anders~S Johansen, Sergio Escalera, Kamal Nasrollahi, Thomas~B
  Moeslund, and Albert Clap{\'e}s.
\newblock Video transformers: A survey.
\newblock {\em IEEE TPAMI}, 2023.

\bibitem{robust}
Madeline~Chantry Schiappa, Naman Biyani, Prudvi Kamtam, Shruti Vyas, Hamid
  Palangi, Vibhav Vineet, and Yogesh~S Rawat.
\newblock A large-scale robustness analysis of video action recognition models.
\newblock In {\em CVPR}, pages 14698--14708, 2023.

\bibitem{kong2022human}
Yu~Kong and Yun Fu.
\newblock Human action recognition and prediction: A survey.
\newblock {\em IJCV}, 130(5):1366--1401, 2022.

\bibitem{channelseparated}
Du~Tran, Heng Wang, Lorenzo Torresani, and Matt Feiszli.
\newblock Video classification with channel-separated convolutional networks.
\newblock In {\em ICCV}, pages 5552--5561, 2019.

\bibitem{tpn}
Ceyuan Yang, Yinghao Xu, Jianping Shi, Bo~Dai, and Bolei Zhou.
\newblock Temporal pyramid network for action recognition.
\newblock In {\em CVPR}, 2020.

\bibitem{nonlocal}
Xiaolong Wang, Ross Girshick, Abhinav Gupta, and Kaiming He.
\newblock Non-local neural networks.
\newblock In {\em CVPR}, pages 7794--7803, 2018.

\bibitem{moreisless}
Quanfu Fan, Chun-Fu~Richard Chen, Hilde Kuehne, Marco Pistoia, and David Cox.
\newblock More is less: Learning efficient video representations by big-little
  network and depthwise temporal aggregation.
\newblock {\em NeurIPS}, 32, 2019.

\bibitem{tin}
Hao Shao, Shengju Qian, and Yu~Liu.
\newblock Temporal interlacing network.
\newblock In {\em AAAI}, pages 11966--11973, 2020.

\bibitem{tsn}
Limin Wang, Yuanjun Xiong, Zhe Wang, Yu~Qiao, Dahua Lin, Xiaoou Tang, and Luc
  Van~Gool.
\newblock Temporal segment networks: Towards good practices for deep action
  recognition.
\newblock In {\em ECCV}, pages 20--36, 2016.

\bibitem{tea}
Yan Li, Bin Ji, Xintian Shi, Jianguo Zhang, Bin Kang, and Limin Wang.
\newblock Tea: Temporal excitation and aggregation for action recognition.
\newblock In {\em CVPR}, pages 909--918, 2020.

\bibitem{teinet}
Zhaoyang Liu, Donghao Luo, Yabiao Wang, Limin Wang, Ying Tai, Chengjie Wang,
  Jilin Li, Feiyue Huang, and Tong Lu.
\newblock Teinet: Towards an efficient architecture for video recognition.
\newblock In {\em AAAI}, volume~34, pages 11669--11676, 2020.

\bibitem{actionet}
Zhengwei Wang, Qi~She, and Aljosa Smolic.
\newblock Action-net: Multipath excitation for action recognition.
\newblock In {\em CVPR}, pages 13214--13223, 2021.

\bibitem{vivit}
Anurag Arnab, Mostafa Dehghani, Georg Heigold, Chen Sun, Mario Lu{\v{c}}i{\'c},
  and Cordelia Schmid.
\newblock Vivit: A video vision transformer.
\newblock In {\em ICCV}, pages 6836--6846, 2021.

\bibitem{zhao2022alignment}
Yizhou Zhao, Zhenyang Li, Xun Guo, and Yan Lu.
\newblock Alignment-guided temporal attention for video action recognition.
\newblock {\em NeurIPS}, 35:13627--13639, 2022.

\bibitem{yan2022multiview}
Shen Yan, Xuehan Xiong, Anurag Arnab, Zhichao Lu, Mi~Zhang, Chen Sun, and
  Cordelia Schmid.
\newblock Multiview transformers for video recognition.
\newblock In {\em CVPR}, pages 3333--3343, 2022.

\bibitem{yang2022recurring}
Jiewen Yang, Xingbo Dong, Liujun Liu, Chao Zhang, Jiajun Shen, and Dahai Yu.
\newblock Recurring the transformer for video action recognition.
\newblock In {\em CVPR}, pages 14063--14073, 2022.

\bibitem{fan-iclr2022}
Rameswar~Panda Quanfu~Fan, Richard~Chen.
\newblock Can an image classifier suffice for action recognition?
\newblock In {\em ICLR}, 2022.

\bibitem{wu2023bidirectional}
Wenhao Wu, Xiaohan Wang, Haipeng Luo, Jingdong Wang, Yi~Yang, and Wanli Ouyang.
\newblock Bidirectional cross-modal knowledge exploration for video recognition
  with pre-trained vision-language models.
\newblock In {\em CVPR}, pages 6620--6630, 2023.

\bibitem{ssv2}
Raghav Goyal, Samira Ebrahimi~Kahou, Vincent Michalski, Joanna Materzynska,
  Susanne Westphal, Heuna Kim, Valentin Haenel, Ingo Fruend, Peter Yianilos,
  Moritz Mueller-Freitag, et~al.
\newblock The something something video database for learning and evaluating
  visual common sense.
\newblock In {\em ICCV}, pages 5842--5850, 2017.

\bibitem{ava}
Chunhui Gu, Chen Sun, David~A Ross, Carl Vondrick, Caroline Pantofaru, Yeqing
  Li, Sudheendra Vijayanarasimhan, George Toderici, Susanna Ricco, Rahul
  Sukthankar, et~al.
\newblock Ava: A video dataset of spatio-temporally localized atomic visual
  actions.
\newblock In {\em CVPR}, pages 6047--6056, 2018.

\bibitem{thumos14}
YG~Jiang, Jingen Liu, A~Roshan Zamir, G~Toderici, I~Laptev, Mubarak Shah, and
  Rahul Sukthankar.
\newblock Thumos challenge: Action recognition with a large number of classe.
\newblock 2014.

\bibitem{imagenet}
Jia Deng, Wei Dong, Richard Socher, Li-Jia Li, Kai Li, and Li~Fei-Fei.
\newblock Imagenet: A large-scale hierarchical image database.
\newblock In {\em CVPR}, pages 248--255, 2009.

\bibitem{afsd}
Chuming Lin, Chengming Xu, Donghao Luo, Yabiao Wang, Ying Tai, Chengjie Wang,
  Jilin Li, Feiyue Huang, and Yanwei Fu.
\newblock Learning salient boundary feature for anchor-free temporal action
  localization.
\newblock In {\em CVPR}, pages 3320--3329, 2021.

\bibitem{mmaction2}
MMAction2 Contributors.
\newblock Openmmlab's next generation video understanding toolbox and
  benchmark.
\newblock \url{https://github.com/open-mmlab/mmaction2}, 2020.

\end{thebibliography}
	
}

\end{document}